\documentclass{article}

\usepackage{arxiv}

\usepackage[utf8]{inputenc} 
\usepackage[T1]{fontenc}    
\usepackage{hyperref}       
\usepackage{url}            
\usepackage{booktabs}       
\usepackage{amsfonts}       
\usepackage{nicefrac}       
\usepackage{microtype}      
\usepackage{subcaption}
\usepackage{tikz}
\usepackage{todonotes}
\usetikzlibrary{shapes,arrows,calc,positioning}

\tikzstyle{line} = [draw, -latex']
\tikzstyle{op} = [rectangle, draw, fill=blue!20, text centered, text width=3em, minimum height=1.5em]
\tikzstyle{logger} = [rectangle, draw, fill=green!20, text centered, text width=3em, minimum height=1.5em]
\tikzstyle{tensor} = [circle, draw, fill=red!20, text centered, text width=1.5em, minimum height=1.5em]

\title{On the quantization of recurrent neural networks}

\author{
  Jian Li\thanks{Correspondence to: jianlijianli@google.com} ~and Raziel Alvarez \\
  Google, Inc., USA.\\
}

\date{}

\renewcommand{\headeright}{}
\renewcommand{\undertitle}{}

\begin{document}
\maketitle

\begin{abstract}
Integer quantization of neural networks can be defined as the approximation of the high precision
computation of the canonical neural network formulation, using reduced integer precision. It plays a significant
role in the efficient deployment and execution of machine learning (ML) systems, reducing memory
consumption and leveraging typically faster computations.
In this work, we present an integer-only quantization strategy for Long Short-Term Memory (LSTM)
neural network topologies, which themselves are the foundation of many production ML systems.
Our quantization strategy is accurate (e.g. works well with quantization post-training), efficient and fast to
execute (utilizing 8 bit integer weights and mostly 8 bit activations), and is able to target a variety of
hardware (by leveraging instructions sets available in common CPU architectures, as well as available neural
accelerators).
\end{abstract}


\section{Introduction}
One important area in the development of machine learning (ML) systems is concerned with the efficient
utilization of the available computational resources. In most practical applications of ML, efficiency
has a direct correlation on whether the system can be put in production at all, due to implications
around quality and cost.

Quantization of neural networks transforms the neural network model's computation such that it can be represented and
executed in a lower precision.
It is an important technique that reduces the memory consumption
(on disk and on RAM), and can help speed up the execution and reduce power consumption \cite{gruenstein2017}.
Moreover, such transformation may be required to execute in a variety of hardware that only supports
integer computations, from digital signal processors \cite{hexagon} to the latest neural accelerators
\cite{edgetpu}.

However, by performing an approximation to a reduced precision form, quantization
may negatively affect the model's quality (i.e. the target loss the neural network was trained to reduce).
Hence, devising quantization strategies that preserve the quality of the original neural networks, while
providing the benefits described above, is a challenging but important area of research.

Such quantization strategies involve defining how the original computation will be approximated,
including decisions about how the original floating point parameters will be converter and represented
in the integer domain, the precision used to represent the resulting outputs from the now
integer computations, and how such resulting intermediate outputs will be converted to a reduced
precision for subsequent computations to take place efficiently. All this while preserving the quality
of the original neural network model.

Whereas significant work has been done in the field over the past several years
\cite{vanhoucke11,han2015learning,alvarez16,jacob17}, the fully integer quantization strategy for
recurrent neural networks (RNNs), and in particular for long short-term memory (LSTM) topologies has always been
challenging.

Recurrent neural networks are widely used to solve tasks involving the processing of sequential data.
In particular, the long short-term memory topology is used in a number of state of the art
production systems, like speech recognition \cite{Graves12, sak2014long, sainath2020}, text translation
\cite{wu16, sutskever14} and text-to-speech~\cite{zen2015, wang17}.
However, the stateful nature of RNN makes its quantization numerically challenging since the
quantization error can accumulate in both depth dimension as well as the sequence (e.g. time) dimension.

There are successful approaches to quantize RNNs. For example, \cite{alvarez16} uses a quantization strategy
that quantizes the static parameters (i.e. weights) in the neural network to 8 bit integers but makes use of a dynamic
computation (i.e. at execution time) of the true floating point ranges of the intermediate values to quantize them
with higher precision. Whereas such approach results in good quality \cite{mcgraw2016, he2019streaming, sainath2020},
and can be implemented efficiently in typical CPU architectures (e.g. x86, arm) resulting in significant
performance gains, it still makes use of some floating point operations and thus lacks the hardware portability,
energy efficiency, and ultimate performance potential that integer-only computation can provide.

With this in mind, this work focuses on the quantization of long short-term memory topologies, with the side
goal that by providing a successful quantization strategy for one of the most useful and complex RNNs, we will be
providing a good foundation to develop quantization strategies for arguably simpler topologies.

This work is organized as follows.
In section~\ref{sec:lstm_architecture} we introduce the different variants of the LSTM topology that we will cover
in our quantization strategy.
In section~\ref{sec:strategy} the quantization strategy is described in full detail.
We start with the over all design approach, which is followed by the building blocks and the design considerations.
The end-to-end quantization strategy is then presented.
This is followed by a discussion in section~\ref{sec:qat} on how the proposed strategy's statistics collection can take
place either during training, or entirely post-training (with a much smaller sub-set of data and without incurring in
fine tuning of the neural network parameters) with good results as validated in the following section.
Experimental results, including accuracy and speed, are listed in section~\ref{sec:result}.
Section~\ref{sec:deployment} describes additional optimizations for deploying models in production.
Finally we provide conclusions in section~\ref{sec:conclusion}.

\section{LSTM architecture}
\label{sec:lstm_architecture}

The long short-term memory cell \cite{hochreiter97} is one of the most complex and widely used RNN topologies.
With time it has seen a number of extensions. In this work we group some of the most widely used such as
the addition of peephole connections~\cite{gers03}, the coupling of input and forget gates (CIFG)~\cite{greff2016lstm},
the addition of a projection~\cite{sak2014long} layer, and the addition of layer normalization~\cite{ba2016layer}.
Thus, in its more complex form containing all but the CIFG extensions mentioned above, the LSTM cell can be described as:

\begin{eqnarray}
i^t = \sigma (norm(W_i x^t + R_i h^{t-1} + P_i \odot c^{t-1}) \odot L_i + b_i)  \label{eq:i1} \\
f^t = \sigma (norm(W_f x^t + R_f h^{t-1} + P_f \odot c^{t-1}) \odot L_f + b_f)  \label{eq:f1} \\
z^t = g(norm(W_z x^t + R_z h^{t-1}) \odot L_z + b_z)    \label{eq:z1} \\
c^t = i^t \odot z^t + f^t \odot c^{t-1}  \label{eq:c1} \\
o^t = \sigma(norm(W_o x^t  + R_o h^{t-1} + P_o \odot c^t) \odot L_o + b_o)  \label{eq:o1} \\
m^t = o^t \odot g(c^t)   \label{eq:m1} \\
h^t = W_{proj} m^t + b_{proj} \label{eq:h1}
\end{eqnarray}

In the LSTM cell definition above (equations~\ref{eq:i1} to~\ref{eq:h1}),
$x^t$ is the input at time $t$, $W$ and $R$ the input
and recurrent weight matrices respectively, $b$ the bias vectors,
$\sigma$ the sigmoid function, $z$, $i$, $f$, $o$ and
$c$ the LSTM-block input, input gate, forget gate, output gate and cell activation vectors,
$h$ the cell output, $\odot$ the element-wise product of the vectors,
and $g$ the activation function, generally $tanh$. We also present other common extensions
such as the peephole connection $P$, the output projection $W_{proj}$ proposed by~\cite{sak2014long} to reduce the
cell size, and the layer normalization computation $norm()$ and terms $L$ added to the input block $z^t$ and gates
($i^t$, $f^t$, and $o^t$) to help stabilize the hidden layer dynamics and speed up model
convergence~\cite{ba2016layer}.

CIFG changes the formulation by "coupling" the input gate to forget gate as $i^{t} = 1 - f^{t}$.
Removing peephole connections, means the term $P \odot c$ is absent.
By removing layer normalization, then the $norm()$ and terms $L$ are absent.
Finally, removing the projection layer implies the hidden state $m$ becomes the output of the cell such that $h = m$.

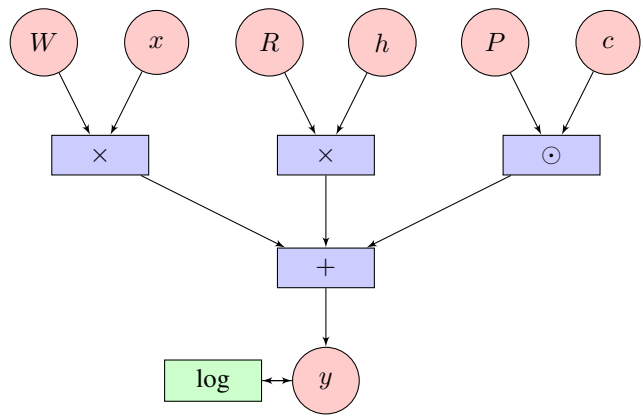
\begin{figure}[t!]
\centering
\begin{tikzpicture}[node distance = 1.5cm, auto]
    \node [tensor] (w_x) {$W$};
    \node [tensor, right of=w_x] (input_x) {$x$};
    \node [tensor, right of=input_x] (w_h) {$R$};
    \node [tensor, right of=w_h] (input_h) {$h$};
    \node [tensor, right of=input_h] (w_cell) {$P$};
    \node [tensor, right of=w_cell] (cell) {$c$};
    \coordinate (x_center) at ($(w_x)!0.5!(input_x)$);
    \coordinate (h_center) at ($(w_h)!0.5!(input_h)$);
    \coordinate (c_center) at ($(w_cell)!0.5!(cell)$);
    \node [op, below of=x_center] (mul_1) {$\times$};
    \node [op, below of=h_center] (mul_2) {$\times$};
    \node [op, below of=c_center] (mul_3) {$\odot$};
    \node [op, below of=mul_2] (add) {$+$};
    \node [tensor, below of=add] (output) {$y$};
    \node [logger, left of=output] (log) {log};

    \path [line] (w_x) -- (mul_1);
    \path [line] (input_x) -- (mul_1);
    \path [line] (w_h) -- (mul_2);
    \path [line] (input_h) -- (mul_2);
    \path [line] (w_cell) -- (mul_3);
    \path [line] (cell) -- (mul_3);
    \path [line] (mul_1) -- (add);
    \path [line] (mul_2) -- (add);
    \path [line] (mul_3) -- (add);
    \path [line] (add) -- (output);
    \path [line] (output) -- (log);
    \path [line] (log) -- (output);
\end{tikzpicture}
\caption{Gates calculation with layer normalization in original float graph.
Cell gate does not have $P$ and $c$.
Note the bias is absent compared with the case with layer normalization.
The output need to be logged.
}
\label{fig:gate_ln_float}
\end{figure}

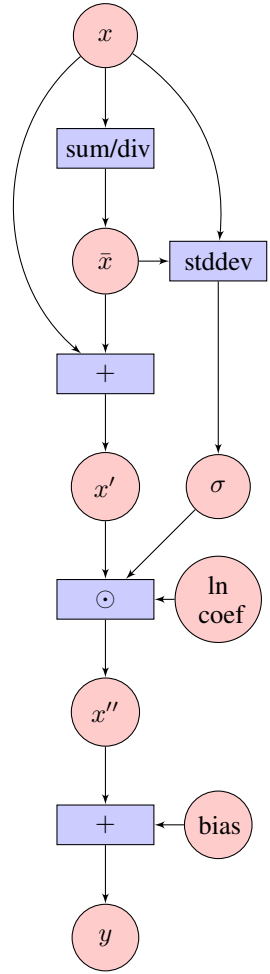
\begin{figure}[t!]
\centering
\begin{tikzpicture}[node distance = 1.5cm, auto]
    \node [tensor] (x) {$x$};
    \node [op, below of=x] (sum) {sum/div};
    \node [tensor, below of=sum] (x_bar) {$\bar{x}$};
    \node [op, right of=x_bar] (stddev) {stddev};
    \node [op, below of=x_bar] (add1) {$+$};
    \node [tensor, below of=add1] (x_1) {$x'$};
    \node [tensor, right of=x_1] (sigma) {$\sigma$};
    \node [op, below of=x_1] (mul1) {$\odot$};
    \node [tensor, right of=mul1] (ln_coef) {ln coef};
    \node [tensor, below of=mul1] (x_3) {$x''$};
    \node [op, below of=x_3] (add2) {$+$};
    \node [tensor, right of=add2] (bias) {bias};
    \node [tensor, below of=add2] (output) {$y$};

    \path [line] (x) -- (sum);
    \path [line] (sum) -- (x_bar);
    \path [line] (x) edge [bend left=30] (stddev);
    \path [line] (x_bar) -- (add1);
    \path [line] (x_bar) -- (stddev);
    \path [line] (stddev) -- (sigma);
    \path [line] (x) edge [bend right=50] (add1);
    \path [line] (add1) -- (x_1);
    \path [line] (x_1) -- (mul1);
    \path [line] (sigma) -- (mul1);
    \path [line] (ln_coef) -- (mul1);
    \path [line] (mul1) -- (x_3);
    \path [line] (x_3) -- (add2);
    \path [line] (bias) -- (add2);
    \path [line] (add2) -- (output);
\end{tikzpicture}
\caption{Layer normalization float calculation.}
\label{fig:ln_float}
\end{figure}

\begin{figure}[t!]
\centering
\begin{tikzpicture}[node distance = 1.5cm, auto]
    \node [tensor] (i_g) {$i$};
    \node [tensor, right of=i_g] (u_g) {$z$};
    \node [tensor, right of=u_g] (f_g) {$f$};
    \node [tensor, right of=f_g] (cell) {$c$};
    \coordinate (center_1) at ($(i_g)!0.5!(u_g)$);
    \coordinate (center_2) at ($(f_g)!0.5!(cell)$);
    \node [op, below of=center_1] (mul1) {$\odot$};
    \node [op, below of=center_2] (mul2) {$\odot$};
    \coordinate (center_3) at ($(mul1)!0.5!(mul2)$);
    \node [op, below of=center_3] (add_1) {$+$};
    \node [tensor, below of=add_1] (cell_2) {$c'$};
    \node [op, left of=cell_2] (o_gate) {output gate};
    \node [tensor, below of=o_gate] (g_output) {$o$};
    \node [op, right of=g_output] (mul3) {$\odot$};
    \node [tensor, below of=mul3] (hidden) {$m$};
    \node [logger, right of=hidden] (log) {log};

    \path [line] (i_g) -- (mul1);
    \path [line] (u_g) -- (mul1);
    \path [line] (f_g) -- (mul2);
    \path [line] (cell) -- (mul2);
    \path [line] (mul1) -- (add_1);
    \path [line] (mul2) -- (add_1);
    \path [line] (add_1) -- (cell_2);
    \path [line] (cell_2) -- (o_gate);
    \path [line] (o_gate) -- (g_output);
    \path [line] (g_output) -- (mul3);
    \path [line] (cell_2) -- (mul3);
    \path [line] (cell_2) edge [bend right=50] (cell);
    \path [line] (mul3) -- (hidden);
    \path [line] (log) -- (hidden);
    \path [line] (hidden) -- (log);

\end{tikzpicture}
\caption{Float calculation from gate to hidden state.
When there is projection, the hidden state needs to be logged.
}
\label{fig:gate_to_hidden_float}
\end{figure}
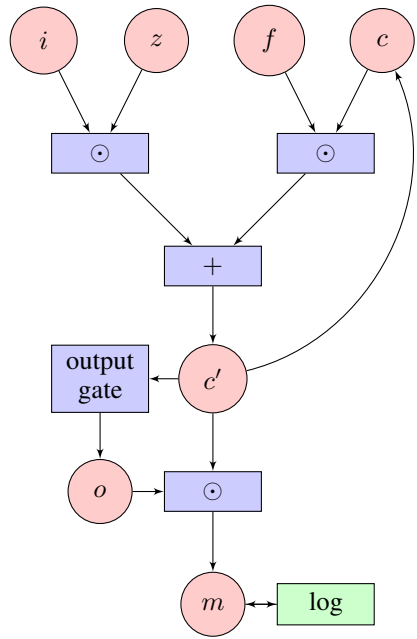

For clarity, we make use of graph diagrams to represent core formulations from the LSTM that will be relevant to
our quantization strategy.
The common gate calculation $W_\lambda x^t + R_\lambda h^{t-1} + P_\lambda \odot c^{t-1}$ (where $\lambda$ can be any
of the gates $i, f, o$) is shown in fig~\ref{fig:gate_no_ln_float} (without
layer normalization) and fig~\ref{fig:gate_ln_float} (with layer normalization).
The layer normalization calculation $norm() + b$ is shown in fig~\ref{fig:ln_float}.
The calculation from the gates to cell state $c^t = i^t \odot z^t + f^t \odot c^{t-1}$, and from the cell to hidden state
$c^t = i^t \odot z^t + f^t \odot c^{t-1}$ is shown in fig~\ref{fig:gate_to_hidden_float}.
The projection layer calculation $h^t = W_{proj} m^t + b_{proj}i$ is shown in fig~\ref{fig:projection_float}.

\section{LSTM quantization strategy}
\label{sec:strategy}

This section lists the details of our quantization strategy for LSTMs, which covers the transformation of floating point
values to integer, as well as the necessary rewrite of the computation for the execution to take place entirely in the integer
domain.

More specifically, our strategy is built around these principles:
\begin{itemize}
\item No floating-point arithmetic: no operation takes place in floating-point, and thus no on-the-fly quantization/de-quantization.
\item No inner loop branching: deeper instruction pipelining.
\item No lookup tables: leverages modern SIMD instruction.
\end{itemize}

More specifically, we aim to leverage 8 bit integer computation as much as possible, and only goes to higher bits when needed.
While existing works have shown that 8 bit quantization is enough for convolution \cite{jacob17}, we have found that is not
enough for accurately approximating the computation in LSTMs without significant accuracy losses.
As such, we start by quantizing all static weight parameters in the LSTM to an 8 bit integer representation, and use
a higher number of bits for calculations that are scalar or non-linear. The relevant details will be covered in the
sections below.

\subsection{Quantization fundamentals}

We quantize values as a linear affine transformation that in principle is similar to many previous
works like \cite{vanhoucke11, alvarez16, jacob17} --to mention a few. This means that at a high level, we quantize each of the
values from a given tensor $T$ of static values (in the original high precision) by computing a linear scale $s_T$ that aims to
evenly distribute them in the narrower scale dictated by the target precision's number of bits $n$ (Equation \ref{basic_scale}).
Thus, each quantized value of the resulting tensor $T_i^{'}$ is transformed according to $s_T$
(Equation \ref{basic_quantized_value}).

\begin{eqnarray}
s_T = \frac{max(T)-min(T)}{2^n} \label{basic_scale} \\
T_i^{'} = \frac{T_i}{s_T} \label{basic_quantized_value}
\end{eqnarray}

Our quantization strategy uses this formulation as its foundation, but important details and decisions around
overflow and saturation, as well as the definition of scales, are described in the following subsections.

\subsubsection{Overflow and saturation}
Matrix multiplication ("matmul") is a basic operation in neural networks, including the LSTM.
It can be defined as a binary operation on two input matrices to produce a third one such that $c_(i_k)=a_{ij}b_{jk}$
where $j$ is accumulated over for all possible values of $i$ and $k$.
Relevant to our work is that its overflow behavior can be modeled via a random walk, and a safe accumulation depth
can be calculated.
For example, in a typical "matmul" operation with inputs comprised of signed 8 bit integers (int8) that accumulate into
a resulting signed 32 bit integer (int32), there is no possibility of overflowing in $2^{15}$ steps. However,
a 24 bit accumulator has only a safe accumulation depth to $2^7$.

Thus, so as long as the dimension of the tensors in the model are smaller than that upper limit of steps, then the
"matmul" computations used to execute the model are safe from accumulator overflowing.
Empirically we have found that, with constraints of "matmul" operations on int8 values to a int32 accumulator,
most models~\cite{alvarez16, shangguan, jacob17} are safe from overflow.

When overflows do occur, in some cases, saturation can solve the overflow issue.
For example, when the operation is followed by a sigmoid function, clamping error due to saturation is
negligibly small outside a proper range such as $[-8, 8]$ (see section~\ref{sec:non-linear}).
However, in other cases, overflow can harm accuracy and should be avoided. Common ways to avoid overflow
(though not entirely guaranteed) are using quantization simulation during training~\cite{jacob17, mot-qat}, or
by selecting a good calibration dataset when post-training quantization is used~\cite{mot-pt}.

The main take away we want to convey here is that overflow and saturation behaviour needs to be modeled carefully, and
thus is a key driver of the decisions (described later in the document) in our proposed quantization strategy.

\subsubsection{Power-of-two scales and $Q_{m.n}$ format}

While in Equation \ref{basic_scale} we define the scale as a real number, in practice we approximate them
as rational numbers defined as power-of-two scales to operate in the fixed-point domain, and because computation
involving power-of-two values can be more efficiently implemented in digital computers as bitwise operations.

We make use of the $Q_{m.n}$ number format \cite{ti2003tms320c64x}, where the integer numbers $m$ and $n$
represent the number of integer and fractional bits, respectively.
More specifically, in this work $Q$ itself denotes signed fixed-point numbers and follows the conversion that $m + n + 1$
equals the bit-width of the type.

This means that $Q_{m.n}$ can represent floating-point values within range of [$-(2^{m})$, $2^{m}-2^{-n}$],
with a resolution of $2^{-n}$.

\subsection{LSTM quantization breakdown}

In this subsection, we breakdown the quantization of common building blocks of the LSTM computation.
This is particularly useful as some times specific decisions need to be made for such blocks, and because
different LSTM variants are constructed from combinations of those blocks.

The LSTM is comprised by a set of gate computations. Each gate computation contains a series of matrix multiplications,
followed by non-linear activation functions ($g$ and $\sigma$) applied to the product of these matrix multiplications.
Even though activation functions of $\sigma$ and $g$ (see eq~\ref{eq:z1}) are in the middle of the LSTM inferencce,
their quantization are discussed first (section~\ref{sec:non-linear}) because they require special input output scales.
Cell state $c$ (see eq~\ref{eq:c1}) is the input to activations so its quantization is determined next in section~\ref{sec:cell-state}.
This is followed by quantization of peephole weight in section~\ref{sec:peephole}, since peephole weight is only used together with cell state $c$.
Gate calculations with layer noramlization ($W x^t + R h^{t-1} + P \odot c^{t-1}$) and without layer noramlization ($W x^t + R h^{t-1} + P \odot c^{t-1} + b$) are discussed in section~\ref{sec:gate-without-ln} and section~\ref{sec:gate-with-ln}.
The quantization of layer normalization is presented in section~\ref{sec:ln}
where we introduced an extra factor in the inference proccess to avoid catastropic accuracy loss that cannot be solved by tuning scales.
The quantization of hidden state (equation~\ref{eq:m1}) is dicussed in section~\ref{sec:gate-to-hidden}.
And when there is projection, the quantization of projection weights and output state (equation~\ref{eq:h1}) is discussed in section~\ref{sec:projection}
At last we discuss the CIFG in section~\ref{sec:cifg} where equation~\ref{eq:i1} becomes $1-f^{t}$.

\subsubsection{Non-linear activation functions} \label{sec:non-linear}
Non-linear activation functions $\sigma$ (see eq~\ref{eq:i1}, eq~\ref{eq:f1}, eq~\ref{eq:o1}) and $g$ (see eq~\ref{eq:c1}) have restrictions on input and output scales
so their scales need to be decided first.

For the activation functions, we utilize 16 bits since we found that given their scalar nature we need
the higher precision to ensure the overall accuracy of the LSTM across different types of models and tasks.
Additionally, $Q_{m.15-m}$ is used as the input and output scale for activations.

For the output of activation functions, $Q_{0,15}$ is selected since it maps nicely to the output range of $\sigma$ and $tanh$.
The output values are slightly clamped at $[-1, \frac{32767}{32768}]$ and experimentally no accuracy loss is observed (see section~\ref{sec:result} for more details).

There are clamping errors and resolutions for activations.
The max of clamping error is $f(\infty) - f(2^{m})$.
Take $tanh$ as an example, if we restrict the input of $tanh$ to $[-8, 8]$ ($Q_{3.12}$), there is a clamping error of $1 - tanh(8) = 2.35e-7$.
Resolution error is the error from representing all values in the quantization "bucket" with one quantized value.
The maximum value of resolution error is $2^{-n}max(f'(x))$, where $max(f'(x))$ is the max value of the derivitive of $f$.
Take $tanh$ as an example, the max resolution error happens at $x = 0$ (max gradient) with values $tanh(2^{-12}) = 2.44e-4$.

When $m$ becomes bigger, clamping error is reduced but resolution error is increased and becomes dominating;
when $m$ becomes smaller, clamping error is dominating.
The value $m$ in $Q_{m.15-m}$ for input of activations is determined by balancing the two errors.
Working out the math for both $tanh$ and $sigma$, $Q_{3.12}$ has the lowest error.

\subsubsection{Cell state} \label{sec:cell-state}
Cell state $c$ is the internal memory of LSTM cell.
Because it persisits accross multiple invocations and mainly involves scalar operations,
16 bits is needed to preserve accuracy.

Cell state is used in 3 places:
\begin{itemize}
  \item as input to elementwise $tanh$.
  \item as input to elementwise multiplication with forget gate.
  \item as input to elementwise multiplication for peephole connection.
\end{itemize}
We cannot clamp cell state to $[-8, 8]$ because it is also used in multiplications.
Instead, we extend the measured range to the next power-of-two values and symmetrically quantize it to 16 bits.
For example, suppose cell is measured to be in the range of $[-3.2, 10]$, we extend that to $[-16, 16)$,
which leads to quantization with $Q_{4.11}$.
And in this case,
even though $Q_{3.12}$ is still the best input format for $tanh$,
we can directly use $Q_{4.11}$ as input to $tanh$ to remove the unnecessary rescaling.

\subsubsection{Peephole connection} \label{sec:peephole}

Peephole connection $P \odot c^{t-1}$ is the optional connection from cell state to gates (see fig~\ref{fig:gate_no_ln_float}).
Since the multiplication between the 16 bit state and the peephole coefficients are elementwise,
we symmetrically quantize the peephole coefficients to $[-32767, 32767]$ with scale $s = \frac{range(max(P))}{32767}$ and store it as int16.
The lack of 16bit-8bit multiplication instruction in ARM neon further deminish the need to quantize peephole weights to 8 bits.

\subsubsection{Gate without layer normalization} \label{sec:gate-without-ln}
There are four gates in one LSTM cell: input (eq~\ref{eq:i1}), forget(eq~\ref{eq:f1}), update(eq~\ref{eq:z1}) and output(eq~\ref{eq:o1}).
Without layer normalization, gate calculation is $W x^t + R h^{t-1} + P \odot c^{t-1} + b$.
As disucssed above, $P$ and $c^{t-1}$ are 16 bits.
For the matrix multiplication operations, 8 bit values and operations are sufficient because the quantization errors are expected to statistically cancel during the accumulation.
$W$ and $R$ are symmetrically quantized to [-127, 127] with scale $s_W = \frac{max(abs())}{127}$ and stored as int8.
$x$ and $h$ are asymmetrically quantized to [-128, 127] with scale $s_x = \frac{max() - min()}{255}$ and stored as int8.
In practice, to ensure float zero point is mapped to an integer, $max(x)$ and $min(x)$ are lightly nudged~\cite{jacob17} in the asymmetric case.

$W x^t$ and $R h^{t-1}$ are accumulations over the product of two int8 so int32 is used to accumulate the result.
For $P \odot c^{t-1} + b$, even though there is no accmulation, int32 is needed for the product of two int16.
So the results of $W x$, $R h$ and $P \odot c$ are stored using three 32 bit accumulators and they have different scales.

Bias is elementwise added to the accumulator so we choose to qunatize that to 32 bits.
Quantized bias needed to be added before rescaling and it can use scale of any of the 3 accumulators.
In the models we tested on, $s_R s_h$ is the smallest out of $s_W s_x$,  $s_R s_h$ and $s_P s_c$ so it provides the best resolution.
So bias is symmetrically quantized with scale $s_R s_h$ to $[-(2^{31}-1), 2^{31}-1]$ and stored as int32.

As discussed in the activation~\ref{sec:non-linear}, without layer normalization,
the overall $W x^t + R h^{t-1} + P \odot c^{t-1} + b$ uses $Q_{3.12}$ as scales so it's symmetrically quantized with scale $s = 2^{-12}$ to $[-32768, 32767]$ and stored as int16.

Figure~\ref{fig:gate_no_ln_quant} shows the quantization details of the gates.

The three int32 accmulators need to be rescaled to the output scale $s = 2^{-12}$ before being added together.
To rescale the accumulators, an effective scale is calculated $s_{effx} = 2^{12} s_{W} s_{x}$ which is the ratio between the accmulator scale and the output scale.
Similarly $s_{effh} = 2^{12} s_{R} s_{h}$ and $s_{effc} = 2^{12} s_{P} s_{c}$ can be calculated.

\begin{figure}[t!]
\centering
\begin{tikzpicture}[node distance = 1.5cm, auto]
    \node [tensor] (w_x) {8};
    \node [tensor, right of=w_x] (input_x) {8,8};
    \node [tensor, right of=input_x] (w_h) {8};
    \node [tensor, right of=w_h] (input_h) {8,8};
    \node [tensor, right of=input_h] (bias) {32};
    \node [tensor, right of=bias] (w_c) {16};
    \node [tensor, right of=w_c] (cell) {$Q_{m.15-m}$};
    \coordinate (x_center) at ($(w_x)!0.5!(input_x)$);
    \coordinate (h_center) at ($(w_h)!0.5!(input_h)$);
    \coordinate (c_center) at ($(w_c)!0.5!(cell)$);
    \node [op, below of=x_center] (mul_1) {$\times$};
    \node [op, below of=h_center] (mul_2) {$\times$};
    \node [op, below of=c_center] (mul_3) {$\odot$};
    \node [tensor, below of=mul_1] (mul_res1) {32};
    \node [tensor, below of=mul_2] (mul_res2) {32};
    \node [tensor, below of=mul_3] (mul_res3) {32};
    \node [op, below of=mul_res2] (add1) {$+$};
    \node [tensor, below of=add1] (add_res1) {32};
    \node [op, below of=add_res1] (rs_2) {rescale};
    \node [op, left of=rs_2] (rs_1) {rescale};
    \node [op, right of=rs_2] (rs_3) {rescale};
    \node [tensor, below of=rs_2] (rs_res2) {$Q_{3.12}$};
    \node [tensor, left of=rs_res2] (rs_res1) {$Q_{3.12}$};
    \node [tensor, right of=rs_res2] (rs_res3) {$Q_{3.12}$};
    \node [op, below of=rs_res2] (add2) {$+$};
    \node [tensor, below of=add2] (output) {$Q_{3.12}$};

    \path [line] (w_x) -- (mul_1);
    \path [line] (input_x) -- (mul_1);
    \path [line] (w_h) -- (mul_2);
    \path [line] (input_h) -- (mul_2);
    \path [line] (w_c) -- (mul_3);
    \path [line] (cell) -- (mul_3);
    \path [line] (mul_1) -- (mul_res1);
    \path [line] (mul_2) -- (mul_res2);
    \path [line] (mul_3) -- (mul_res3);
    \path [line] (mul_res1) -- (rs_1);
    \path [line] (mul_res3) -- (rs_3);
    \path [line] (mul_res2) -- (add1);
    \path [line] (bias) edge [bend left=30] (add1);
    \path [line] (rs_1) -- (rs_res1);
    \path [line] (rs_3) -- (rs_res3);
    \path [line] (add1) -- (add_res1);
    \path [line] (add_res1) -- (rs_2);
    \path [line] (rs_2) -- (rs_res2);
    \path [line] (rs_res1) -- (add2);
    \path [line] (rs_res2) -- (add2);
    \path [line] (rs_res3) -- (add2);
    \path [line] (add2) -- (output);
\end{tikzpicture}
\caption{Integer execution for gates calculation without layer normalization. Cell gate does not have $P$ and $c$. The multiplication between (8) and (8, 8) is w (x + zp), which in practice is handled as w x + w zp.}
\label{fig:gate_no_ln_execution}
\end{figure}
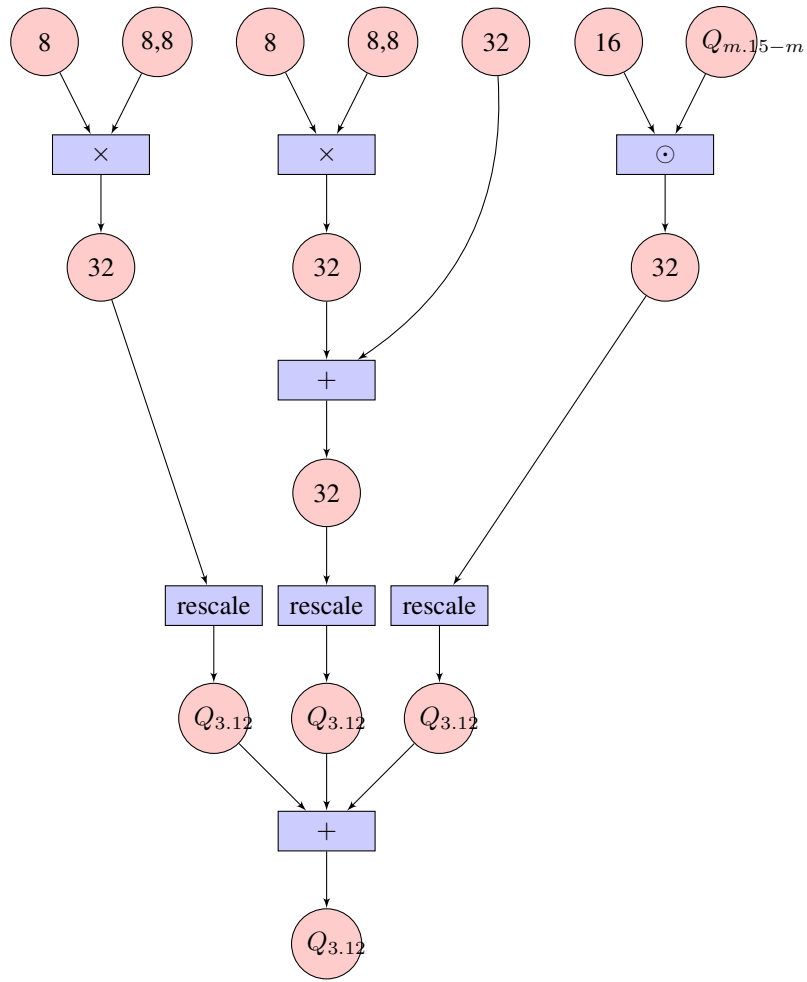

The gate integer execution becomes $\textsf{rescale}(W (x + zp), s_{effx}) + \textsf{rescale}(R (h+zp) + b, s_{effh}) + \textsf{rescale}(Pc, s_{effc})$.
This is visualized in figure~\ref{fig:gate_no_ln_execution}.

\subsubsection{Gate with layer normalization} \label{sec:gate-with-ln}

In the presence of layer normalizatoin, the gate calculation becomes $W x + R h + P \odot c$ and the bias is applied after normalization.
Same as the cases without layer normalization $W$, $x$, $R$, $h$ are quantized to 8 bit and $P$ and $c$ are quantized to 16 bit.
Because the output of the gate is not consumed by activations, we cannot use $2^{-12}$ as the output scale.
Instead the output scale is calculated from measured values: $s = \frac{max(abs(W x + R h + P \odot c))}{32767}$.
The quantized values are mapped to [-32767, 32767] and stored as int16.
The quantization is shown in figure~\ref{fig:gate_ln_quant}.

\begin{figure}[t!]
\centering
\begin{tikzpicture}[node distance = 1.5cm, auto]
    \node [tensor] (w_x) {8};
    \node [tensor, right of=w_x] (input_x) {8,8};
    \node [tensor, right of=input_x] (w_h) {8};
    \node [tensor, right of=w_h] (input_h) {8,8};
    \node [tensor, right of=input_h] (w_c) {16};
    \node [tensor, right of=w_c] (cell) {$Q_{m.15-m}$};
    \coordinate (x_center) at ($(w_x)!0.5!(input_x)$);
    \coordinate (h_center) at ($(w_h)!0.5!(input_h)$);
    \coordinate (c_center) at ($(w_c)!0.5!(cell)$);
    \node [op, below of=x_center] (mul_1) {$\times$};
    \node [op, below of=h_center] (mul_2) {$\times$};
    \node [op, below of=c_center] (mul_3) {$\odot$};
    \node [tensor, below of=mul_1] (mul_res1) {32};
    \node [tensor, below of=mul_2] (mul_res2) {32};
    \node [tensor, below of=mul_3] (mul_res3) {32};
    \node [op, below of=mul_res1] (rs_1) {rescale};
    \node [op, below of=mul_res2] (rs_2) {rescale};
    \node [op, below of=mul_res3] (rs_3) {rescale};
    \node [tensor, below of=rs_1] (rs_res1) {16};
    \node [tensor, below of=rs_2] (rs_res2) {16};
    \node [tensor, below of=rs_3] (rs_res3) {16};
    \node [op, below of=rs_res2] (add2) {$+$};
    \node [tensor, below of=add2] (output) {16};

    \path [line] (w_x) -- (mul_1);
    \path [line] (input_x) -- (mul_1);
    \path [line] (w_h) -- (mul_2);
    \path [line] (input_h) -- (mul_2);
    \path [line] (w_c) -- (mul_3);
    \path [line] (cell) -- (mul_3);
    \path [line] (mul_1) -- (mul_res1);
    \path [line] (mul_2) -- (mul_res2);
    \path [line] (mul_3) -- (mul_res3);
    \path [line] (mul_res1) -- (rs_1);
    \path [line] (mul_res2) -- (rs_2);
    \path [line] (mul_res3) -- (rs_3);
    \path [line] (rs_1) -- (rs_res1);
    \path [line] (rs_2) -- (rs_res2);
    \path [line] (rs_3) -- (rs_res3);
    \path [line] (rs_res1) -- (add2);
    \path [line] (rs_res2) -- (add2);
    \path [line] (rs_res3) -- (add2);
    \path [line] (add2) -- (output);
\end{tikzpicture}
\caption{Integer execution for gates calculation with layer normalization. Cell gate does not have $P$ and $c$. The multiplication between (8) and (8, 8) is w (x + zp), which in practice is handled as w x + w zp.}
\label{fig:gate_ln_execution}
\end{figure}
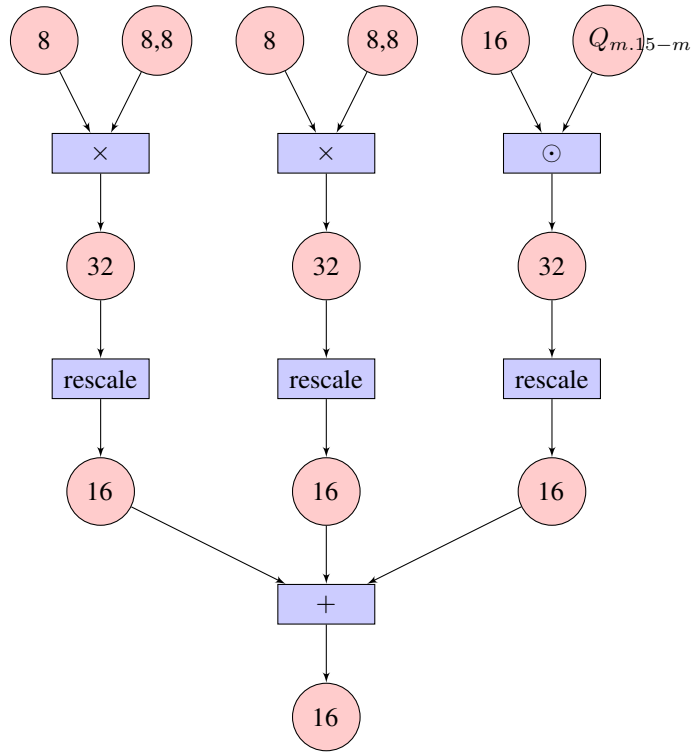

Same as the non layer normalization case, three effective scales ($s_{effx}$, $s_{effh}$, $s_{effc}$) are calculated.
Integer execution is $\textsf{rescale}(W (x + zp), s_{effx}) + \textsf{rescale}(R (h+zp), s_{effh}) + \textsf{rescale}(Pc, s_{effc})$,
which can be visualized in figure~\ref{fig:gate_ln_execution}.

\subsubsection{Layer normalization} \label{sec:ln}

Layer normalization~\cite{ba2016layer} $norm() \odot L + b$ is widely used in streaming use cases.
It normalizes the activation vector therefore prevents the model from overall shifts caused by the gate matrix multiplication,
which is one of the primary source of accuracy degradation for LSTM cell quantization.

Layer normalization normalize $x$ to $x'$ with zero mean and unity standard deviation and then applies layer normalization coefficients and bias.
\begin{eqnarray}
\bar{x} = \frac{\sum_{i = 1}^{n}(x_i)}{n} \label{ln_mean} \\
x_i^{'} = \frac{(x_i - \bar{x})}{\sqrt{\frac{\sum_{i=1}^{n}(x_i^2 - \bar{x}_i^2)}{n}}} \label{ln_norm} \\
y_i = x_i^{'} * L_{i} + b_i \label{ln_dot}
\end{eqnarray}
The float calculation is shown in fig~\ref{fig:ln_float}

The output of normalization $x'$ is limited to a small range.
Assuming normal distribution, 99.7\% of values in $x_i^{'}$ is confined between $[-3.0, 3.0]$ which is roughly 2.8 bits in the integer presentation.
This leads to catastrophic accuracy degradation in the model.
Increasing bits and/or adjusting scales of the input of normalization would not help because any scale would be cancelled since it appears in both the numerator and denominitor in the calculation of $x_i^{'}$.

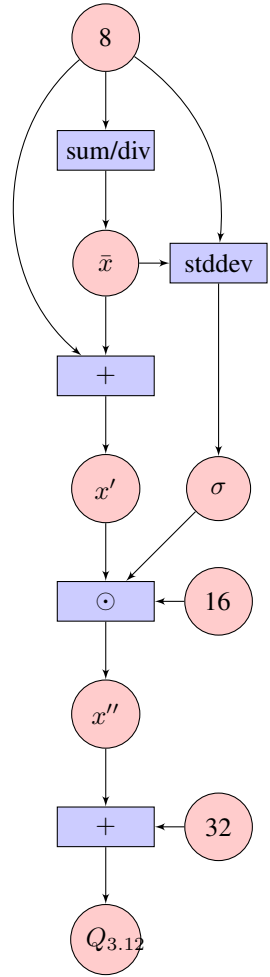
\begin{figure}[t!]
\centering
\begin{tikzpicture}[node distance = 1.5cm, auto]
    \node [tensor] (x) {8};
    \node [op, below of=x] (sum) {sum/div};
    \node [tensor, below of=sum] (x_bar) {$\bar{x}$};
    \node [op, right of=x_bar] (stddev) {stddev};
    \node [op, below of=x_bar] (add1) {$+$};
    \node [tensor, below of=add1] (x_1) {$x'$};
    \node [tensor, right of=x_1] (sigma) {$\sigma$};
    \node [op, below of=x_1] (mul1) {$\odot$};
    \node [tensor, right of=mul1] (ln_coef) {16};
    \node [tensor, below of=mul1] (x_3) {$x''$};
    \node [op, below of=x_3] (add2) {$+$};
    \node [tensor, right of=add2] (bias) {32};
    \node [tensor, below of=add2] (output) {$Q_{3.12}$};

    \path [line] (x) -- (sum);
    \path [line] (sum) -- (x_bar);
    \path [line] (x) edge [bend left=30] (stddev);
    \path [line] (x_bar) -- (add1);
    \path [line] (x_bar) -- (stddev);
    \path [line] (stddev) -- (sigma);
    \path [line] (x) edge [bend right=50] (add1);
    \path [line] (add1) -- (x_1);
    \path [line] (x_1) -- (mul1);
    \path [line] (sigma) -- (mul1);
    \path [line] (ln_coef) -- (mul1);
    \path [line] (mul1) -- (x_3);
    \path [line] (x_3) -- (add2);
    \path [line] (bias) -- (add2);
    \path [line] (add2) -- (output);
\end{tikzpicture}
\caption{Layer normalization in quantized format.} \label{fig:ln_quant}
\end{figure}

Instead, we solve the challenge by adding a scaling factor $s'$ to $x'$ in the computational graph.
With $x_i^{'} = q_i^{'} s'$, the quantized value $q_i^{'}$ can now be expressed as
$q_i^{'} = \frac{x_i^{'}}{s'} = \frac{(q_i - \bar{q})}{\sqrt{\frac{\sum_{i=1}^{n}(q_i^2 - \bar{q}_i^2)}{n}}} \times \frac{1}{s'}$.
$s'$ is chosen to be the smallest power-of-two number that won't cause overflows in $q_i^{'}$, which is $2^{-10}$ in our experiments.
With $s'$ determined, the integer inference becomes~\cite{shangguan}:
\begin{eqnarray}
\bar{q} = round(\frac{\sum_{i = 1}^{n}(2^{10} q_i)}{n}) \\
\sigma = \sqrt{\frac{\sum_{i=1}^{n}(2^{20} q_i^2 - \bar{q}_i^2)}{n}} = \sqrt{\frac{2^{20}}{n} \sum_{i=1}^{n} q_i^2 - \bar{q}_i^2}\\
q_i^{'} = round(\frac{(2^{10} q_i - \bar{q})}{\sigma}) \\
q_i^{''} = round(\frac{q_i^{'} q_{Li} + q_{bi}}{2^{10}})
\end{eqnarray}
$q_{L}$ is the quantized value of weight and $q_{b}$ is quantized value of bias.
Layer normalization weights are symmetrically quantized with scale $s_{L} = \frac{range(L)}{32767}$ to [-32767, 32767] and stored as int16.
Bias is symmetrically quantized with scale $s_b = 2^{-10} s_{L}$ to $[-(2^{31}-1), 2^{31}-1]$ and stored as int32.
The quantization of layer normalization is shown in fig~\ref{fig:ln_quant} and the integer execution is shown in fig~\ref{fig:ln_execution}.

\subsubsection{Gate outputs to hidden} \label{sec:gate-to-hidden}
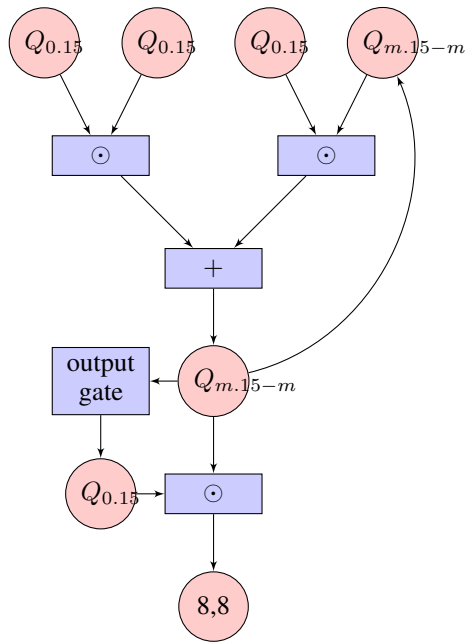
\begin{figure}[t!]
\centering
\begin{tikzpicture}[node distance = 1.5cm, auto]
    \node [tensor] (i_g) {$Q_{0.15}$};
    \node [tensor, right of=i_g] (u_g) {$Q_{0.15}$};
    \node [tensor, right of=u_g] (f_g) {$Q_{0.15}$};
    \node [tensor, right of=f_g] (cell) {$Q_{m.15-m}$};
    \coordinate (center_1) at ($(i_g)!0.5!(u_g)$);
    \coordinate (center_2) at ($(f_g)!0.5!(cell)$);
    \node [op, below of=center_1] (mul1) {$\odot$};
    \node [op, below of=center_2] (mul2) {$\odot$};
    \coordinate (center_3) at ($(mul1)!0.5!(mul2)$);
    \node [op, below of=center_3] (add_1) {$+$};
    \node [tensor, below of=add_1] (cell_2) {$Q_{m.15-m}$};
    \node [op, left of=cell_2] (o_gate) {output gate};
    \node [tensor, below of=o_gate] (g_output) {$Q_{0.15}$};
    \node [op, right of=g_output] (mul3) {$\odot$};
    \node [tensor, below of=mul3] (hidden) {8,8};

    \path [line] (i_g) -- (mul1);
    \path [line] (u_g) -- (mul1);
    \path [line] (f_g) -- (mul2);
    \path [line] (cell) -- (mul2);
    \path [line] (mul1) -- (add_1);
    \path [line] (mul2) -- (add_1);
    \path [line] (add_1) -- (cell_2);
    \path [line] (cell_2) -- (o_gate);
    \path [line] (o_gate) -- (g_output);
    \path [line] (g_output) -- (mul3);
    \path [line] (cell_2) -- (mul3);
    \path [line] (cell_2) edge [bend right=50] (cell);
    \path [line] (mul3) -- (hidden);

\end{tikzpicture}
\caption{From gate to hidden state quantized}
\label{fig:gate_to_hidden_quant}
\end{figure}

The new cell state $c^{t}$ is calculated using the old cell state $c^{t-1}$ and input gate, forget gate and update gate through $c^t = i^t \odot z^t + f^t \odot c^{t-1}$.
As shown in fig~\ref{fig:gate_to_hidden_quant}, the 3 gates all have $Q_{0.15}$ as scale and cell has $Q_{m.15-m}$ as scale.
The quantized execution from gate to cell is $c^t = \textsf{shift}(i^t \odot z^t, 30 - m) + \textsf{shift}(f^t \odot c^{t-1}, 15)$

Hidden state $m^{t} = o^t \odot g(c^t)$ is the element wise product between the output gate and cell state.
Hidden state is mathmatically bounded to $[-1, 1]$ because it is the product of two values that are bounded to $[-1, 1]$.
Experimentally we have observed that using the measured range instead of $[-1, 1]$ improves accuracy.
So the hidden state is quantized asymmetrically with scale $s_{m} = range(m) / 255$ to [-128, 127] and stored as int8.

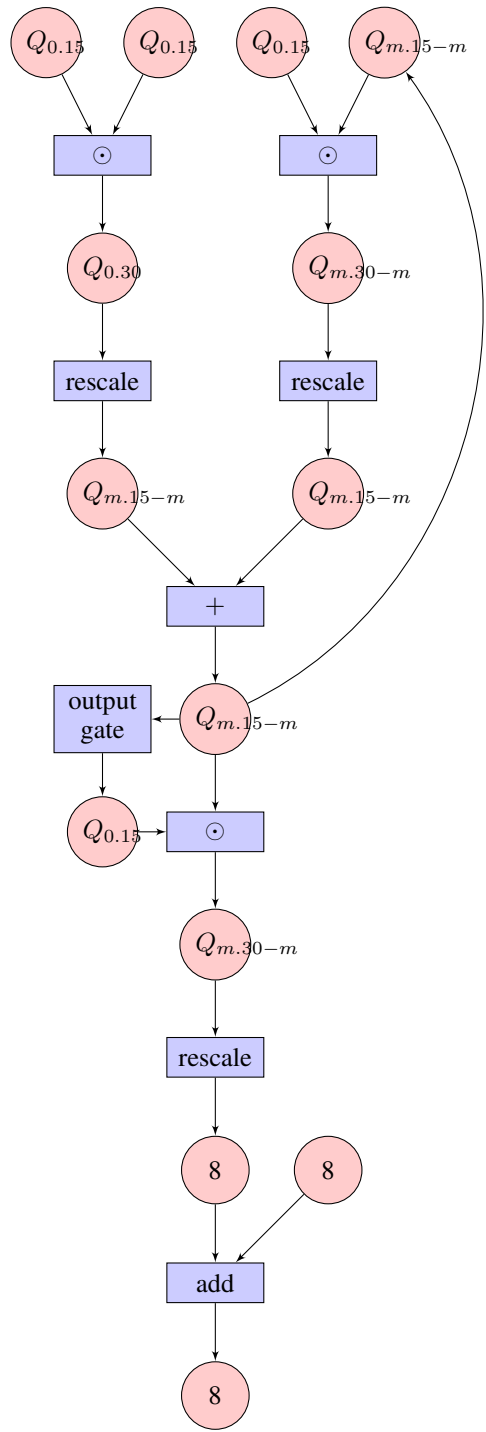
\begin{figure}[t!]
\centering
\begin{tikzpicture}[node distance = 1.5cm, auto]
    \node [tensor] (i_g) {$Q_{0.15}$};
    \node [tensor, right of=i_g] (u_g) {$Q_{0.15}$};
    \node [tensor, right of=u_g] (f_g) {$Q_{0.15}$};
    \node [tensor, right of=f_g] (cell) {$Q_{m.15-m}$};
    \coordinate (center_1) at ($(i_g)!0.5!(u_g)$);
    \coordinate (center_2) at ($(f_g)!0.5!(cell)$);
    \node [op, below of=center_1] (mul1) {$\odot$};
    \node [op, below of=center_2] (mul2) {$\odot$};
    \node [tensor, below of=mul1] (mul_res_1) {$Q_{0.30}$};
    \node [tensor, below of=mul2] (mul_res_2) {$Q_{m.30-m}$};
    \node [op, below of=mul_res_1] (res_0_1) {rescale};
    \node [op, below of=mul_res_2] (res_0_2) {rescale};
    \node [tensor, below of=res_0_1] (mul_res_scaled_1) {$Q_{m.15-m}$};
    \node [tensor, below of=res_0_2] (mul_res_scaled_2) {$Q_{m.15-m}$};
    \coordinate (center_3) at ($(mul_res_scaled_1)!0.5!(mul_res_scaled_2)$);
    \node [op, below of=center_3] (add_1) {$+$};
    \node [tensor, below of=add_1] (cell_2) {$Q_{m.15-m}$};
    \node [op, left of=cell_2] (o_gate) {output gate};
    \node [tensor, below of=o_gate] (g_output) {$Q_{0.15}$};
    \node [op, right of=g_output] (mul3) {$\odot$};
    \node [tensor, below of=mul3] (mul_output) {$Q_{m.30-m}$};
    \node [op, below of=mul_output] (rescale) {rescale};
    \node [tensor, below of=rescale] (temp) {8};
    \node [tensor, right of=temp] (zp) {8};
    \node [op, below of=temp] (add) {add};
    \node [tensor, below of=add] (hidden) {8};

    \path [line] (i_g) -- (mul1);
    \path [line] (u_g) -- (mul1);
    \path [line] (f_g) -- (mul2);
    \path [line] (cell) -- (mul2);
    \path [line] (mul1) -- (mul_res_1);
    \path [line] (mul2) -- (mul_res_2);
    \path [line] (mul_res_1) -- (res_0_1);
    \path [line] (mul_res_2) -- (res_0_2);
    \path [line] (res_0_1) -- (mul_res_scaled_1);
    \path [line] (res_0_2) -- (mul_res_scaled_2);
    \path [line] (mul_res_scaled_1) -- (add_1);
    \path [line] (mul_res_scaled_2) -- (add_1);
    \path [line] (add_1) -- (cell_2);
    \path [line] (cell_2) -- (o_gate);
    \path [line] (o_gate) -- (g_output);
    \path [line] (g_output) -- (mul3);
    \path [line] (cell_2) -- (mul3);
    \path [line] (cell_2) edge [bend right=50] (cell);
    \path [line] (mul3) -- (mul_output);
    \path [line] (mul_output) -- (rescale);
    \path [line] (rescale) -- (temp);
    \path [line] (temp) -- (add);
    \path [line] (zp) -- (add);
    \path [line] (add) -- (hidden);

\end{tikzpicture}
\caption{Integer execution from gate to hidden state.}
\label{fig:gate_to_hidden_execution}
\end{figure}

From cell to hidden state $m$, an effective scale of $s_{eff} = 2^{-30} / s_{m}$ is calculated and the integer execution is
$m = \textsf{rescale}(o \odot g(c), s_{eff}) - zp$.
The overallinteger execution from gate to hidden state is shown in fig~\ref{fig:gate_to_hidden_execution}.

\subsubsection{Projection} \label{sec:projection}

\begin{figure}[t!]
\centering
\begin{tikzpicture}[node distance = 1.5cm, auto]
    \node [tensor] (weight) {8};
    \node [tensor, right of=weight] (input) {$8,8$};
    \node [tensor, right of=input] (bias) {$32$};
    \coordinate (h_center) at ($(weight)!0.5!(input)$);
    \node [op, below of=h_center] (mul) {$\times$};
    \node [op, below of=mul] (add) {$+$};
    \node [tensor, below of=add] (output) {$8$};

    \path [line] (weight) -- (mul);
    \path [line] (input) -- (mul);
    \path [line] (mul) -- (add);
    \path [line] (bias) -- (add);
    \path [line] (add) -- (output);
\end{tikzpicture}
\caption{Projection calulation in quantized format.}
\label{fig:projection_quant}
\end{figure}
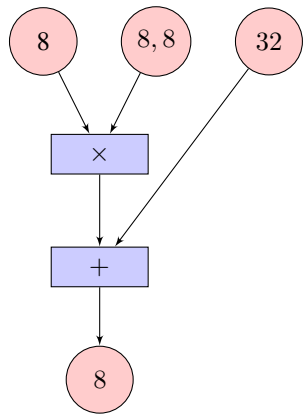

Projection $h^t = W_{proj} m^t + b_{proj}$ is the optional connection that projects the hidden statee $m$ to output $h$.
$W$ is symmetrically quantized with scale $s_w = max(abs(W_{proj}))/127$ to [-127, 127] and stored as int8.
Bias $b_{proj}$ is symmerically quantized with scale $s_b = s_w * s_m$ to $[-(2^{31}-1), 2^{31}-1]$ and stored as int32.
$h$ is asymmetrically quantized with scale $s_h = range(h)/255$ to [-128, 127] and stored as int8.
This is shown in fig~\ref{fig:projection_quant}.

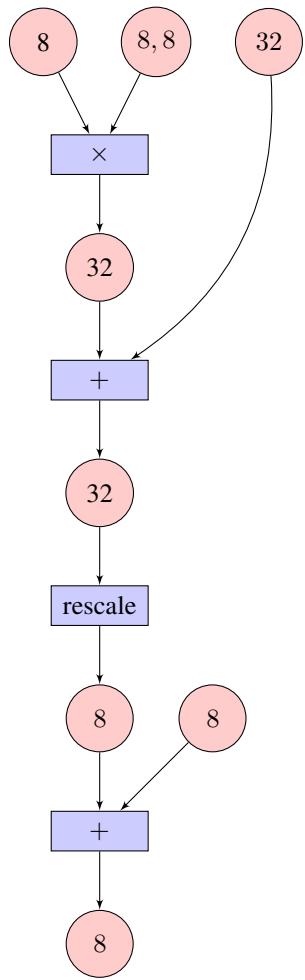
\begin{figure}[t!]
\centering
\begin{tikzpicture}[node distance = 1.5cm, auto]
    \node [tensor] (weight) {8};
    \node [tensor, right of=weight] (input) {$8,8$};
    \node [tensor, right of=input] (bias) {$32$};
    \coordinate (h_center) at ($(weight)!0.5!(input)$);
    \node [op, below of=h_center] (mul) {$\times$};
    \node [tensor, below of=mul] (mul_res) {32};
    \node [op, below of=mul_res] (add) {$+$};
    \node [tensor, below of=add] (add_res) {32};
    \node [op, below of=add_res] (rescale) {rescale};
    \node [tensor, below of=rescale] (rescale_res) {$8$};
    \node [tensor, right of=rescale_res] (zp) {$8$};
    \node [op, below of=rescale_res] (add_2) {$+$};
    \node [tensor, below of=add_2] (output) {$8$};

    \path [line] (weight) -- (mul);
    \path [line] (input) -- (mul);
    \path [line] (mul) -- (mul_res);
    \path [line] (mul_res) -- (add);
    \path [line] (add) -- (add_res);
    \path [line] (bias) edge [bend left=30] (add);
    \path [line] (add_res) -- (rescale);
    \path [line] (rescale) -- (rescale_res);
    \path [line] (rescale_res) -- (add_2);
    \path [line] (zp) -- (add_2);
    \path [line] (add_2) -- (output);
\end{tikzpicture}
\caption{Integer execution for projection. The last $add$ is for zero point of output $h$.}
\label{fig:projection_execution}
\end{figure}

Similarly, with effective scale $s_{eff} = s_{W_{proj}} * s_m / s_h$, the integer execution is $h = \textsf{rescale}(W_{proj} (m + m_{zp}) + b, s_{eff}) - h_{zp}$ and is shown in fig~\ref{fig:projection_execution}.

\subsubsection{CIFG} \label{sec:cifg}

With CIFG, the input gate and forget gate are coupled: $i^{t} = 1 - f^{t}$.
In the quantized case, because input and forget gate have $2^{-15}$ as scale, the coupling becomes $i^{t} = max(32768 - f^{t}, 32767)$.
The extra clamping is to make sure the result can fit in int16.
And becasue the forget gate is clamped slightly to $[0, \frac{32767}{32768}]$ (instead of $[0, 1]$),
input gate is clamped to $[\frac{1}{32768}, \frac{32767}{32768}]$ for CIFG.

\subsection{Quantization strategy}

The quantization of all variants of LSTM can built on top of the above mentioned building blocks.
They are summerized in table~\ref{tab:recipe}.
On the high level, the quantization is:
\begin{itemize}
\item Matrix related operations, such as matrix-matrix multiplication are in 8-bit;
\item Vector related operations, such as element wise sigmoid, are in a mixture of 8-bit and 16-bit.
\end{itemize}

\section{Collecting statistics}
\label{sec:qat}

As shown in equation~\ref{basic_scale}, in order to calculate the scale, maximum and minimum values of the tensor are needed.
For static weights, maximum and minimum values can be easily obtained.
For activations, there are two ways of collecting maximum and minimum values: Post Training~\cite{tfmot} and Quantization Aware Training (QAT)~\cite{jacob17, mot-qat}.
Post Training runs inference on a representative data set and collect statistics;
QAT collects tensor statistics during training and additionally fine tunes model weight by simulating quantization noise in the training process.
Quantization strategy, including both quantization of the model and the integer execution, applies both Post Training and QAT.
Post-training quantization is easier to use because it does not require training.
Currently LSTM quantization is enabled in the post-training approach in TensorFlow.

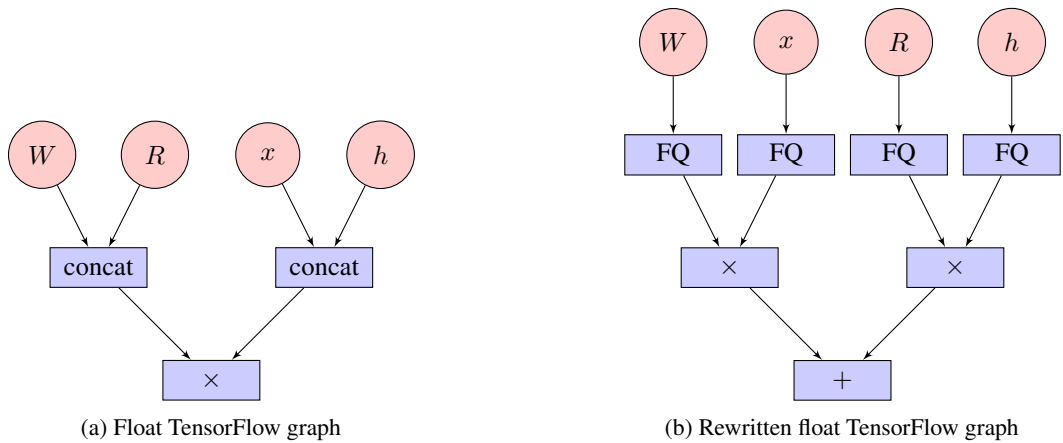
\begin{figure}[t!]
\centering
  \begin{subfigure}[t]{0.5\textwidth}
  \centering
  \begin{tikzpicture}[node distance = 1.5cm, auto]
      \node [tensor] (w_x) {$W$};
      \node [tensor, right of=w_x] (w_h) {$R$};
      \node [tensor, right of=w_h] (input_x) {$x$};
      \node [tensor, right of=input_x] (input_h) {$h$};
      \coordinate (w_x_w_h_center) at ($(w_x)!0.5!(w_h)$);
      \coordinate (x_h_center) at ($(input_x)!0.5!(input_h)$);
      \node [op, below of=w_x_w_h_center] (concat_1) {concat};
      \node [op, below of=x_h_center] (concat_2) {concat};
      \coordinate (all_center) at ($(concat_1)!0.5!(concat_2)$);
      \node [op, below of=all_center] (mul) {$\times$};

      \path [line] (w_x) -- (concat_1);
      \path [line] (w_h) -- (concat_1);
      \path [line] (input_x) -- (concat_2);
      \path [line] (input_h) -- (concat_2);
      \path [line] (concat_1) -- (mul);
      \path [line] (concat_2) -- (mul);
  \end{tikzpicture}
  \caption{Float TensorFlow graph} \label{fig:qat0}
  \end{subfigure}%
~
  \begin{subfigure}[t]{0.5\textwidth}
  \centering
  \begin{tikzpicture}[node distance = 1.5cm, auto]
      \node [tensor] (w_x) {$W$};
      \node [tensor, right of=w_x] (input_x) {$x$};
      \node [tensor, right of=input_x] (w_h) {$R$};
      \node [tensor, right of=w_h] (input_h) {$h$};
      \node [op, below of=w_x] (fq_w_x) {FQ};
      \node [op, below of=input_x] (fq_x) {FQ};
      \node [op, below of=w_h] (fq_w_h) {FQ};
      \node [op, below of=input_h] (fq_h) {FQ};
      \coordinate (w_x_x_center) at ($(fq_w_x)!0.5!(fq_x)$);
      \coordinate (w_h_h_center) at ($(fq_w_h)!0.5!(fq_h)$);
      \node [op, below of=w_x_x_center] (matmul_1) {$\times$};
      \node [op, below of=w_h_h_center] (matmul_2) {$\times$};
      \coordinate (all_center) at ($(matmul_1)!0.5!(matmul_2)$);
      \node [op, below of=all_center] (add) {$+$};

      \path [line] (w_x) -- (fq_w_x);
      \path [line] (input_x) -- (fq_x);
      \path [line] (w_h) -- (fq_w_h);
      \path [line] (input_h) -- (fq_h);

      \path [line] (fq_w_x) -- (matmul_1);
      \path [line] (fq_x) -- (matmul_1);
      \path [line] (fq_w_h) -- (matmul_2);
      \path [line] (fq_h) -- (matmul_2);

      \path [line] (matmul_1) -- (add);
      \path [line] (matmul_2) -- (add);
  \end{tikzpicture}
  \caption{Rewritten float TensorFlow graph} \label{fig:qat1}
  \end{subfigure}
\caption{
Float graph modification for quantization aware training of LSTM.
Input activation and recurrent activations have different scales so fake quant need to added to each components with concatination.
}
\label{fig:qat}
\end{figure}

For the sake of completeness, we document the QAT approach for LSTM as well.
In the original TensorFlow graph, for performance reasons,
the recurrent and input activations are concatenated and the recurrent weight and input activation weights are concatenated.
Since the weights and activations in integer LSTM have saparate scales for input and recurrent calculation,
the training graph need to be rewritten to remove the concatenation of the weights and activations for all the gates.
The graph before and after the rewriting are shown in fig~\ref{fig:qat}.

\section{Experiments}
\label{sec:result}
In table~\ref{tab:benchmark}, we reproduce~\cite{shangguan} the accuracy for speech recognization on 3 anonymized private benchmark speech datasets on VoiceSearch, YouTube and Telephony.
VoiceSearch and Telephony dataset have average utterance length of 4.7 seconds while YouTube dataset contains longer utterances, averaging 16.5 min per utterance.
All models have the same RNN Transducer (RNN-T)\cite{Graves12, he2019streaming} architeture with 10 layers of LSTM and each of them contains 2048 hidden units.
A fixed 100-utterances dataset is sufficient to quantize the model with negligible accuracy loss across all LSTM architecture and sparsity levels,
despite the large variety of audio and semantic conditions in speech recognition.
The good accuracy on the long utternaces in YouTube dataset shows the robustness of the quantizaiton strategy.

\begin{table*}[ht!]
\centering
\begin{tabular}{|l|l|l|l|l|c|l|l|} \hline
              & Enc\&Dec & Sparsity  &\#Params(M)  & Quantization            & \multicolumn{3}{c|}{WER on dataset}  \\\hline
              &          &           &\% baseline  & Size(MB)                & VoiceSearch  & YouTube & Telephony         \\\hline
LSTM          & LSTMx8   & 0\%       & 122.1       & Float,466MB            & 6.6          & 19.5    & 8.1       \\
(baseline)    & LSTMx2   & 0\%       & 100\%       & Hybrid,117MB           & 6.7          & 19.8    & 8.2        \\
              &          &           &             & Integer,117MB          & 6.7          & 19.8    & 8.2          \\
\hline
Sparse LSTM   & LSTMx8   & 50\%      & 69.7        & Float,270MB            & 6.7          & 20.2    & 8.2        \\
              & LSTMx2   & 50\%      & 57\%        & Hybrid,71MB            & 6.8          & 20.4    & 8.4          \\
              &          &           &             & Integer, 71MB           & 6.9          & 22.9    & 8.7         \\
\hline
Sparse CIFG   & CIFGx8   & 50\%      & 56.3        & Float,219MB            & 7.1          & 21.7    & 8.3      \\
              & CIFGx2   & 50\%      & 46\%        & Hybrid,57MB            & 7.2          & 21.4    & 8.5       \\
              &          &           &             & Integer,57MB           & 7.2          & 20.6    & 8.7       \\
\hline
\end{tabular}
\caption{Comparison of float, hybrid and fully quantized models across different dataset. Table is reproduced from~\cite{shangguan}}
\label{tab:benchmark}
\end{table*}

\section{Deployment}
\label{sec:deployment}

In this section we discuss an optimization for integer execution of LSTM that are improtant for deployment.

The most computational heavy part of the execution is the accumulation on the product of weight and activation:
$\Sigma_i{(W_i \times (x_i - zp))} + b_i$.
The calculation on zero point $\Sigma_i{(W_i} \times zp)$ can be performed offline since they are static,
In TensorFlow Lite\cite{tflite}, the pre-calculated zero points are folded into the bias
$\Sigma_i{(W_i \times x_i)} + \Sigma_i{(W_i \times zp)} + b_i = \Sigma_i{(W_i \times x_i)} + b^{'}_i$
and the acutal kernel treats both weight and activation symmetric.
With this optimization, the integer LSTM is about ~5\% faster than hybrid and two times faster than float in RT factor~\cite{shangguan}.

\section{Conclusions}
\label{sec:conclusion}
We have demonstrated that integer RNN is accurate and meaningfully faster on CPU.
More importantly, integer quantization has a number of additional advantages:
\begin{itemize}
\item Widespread availability. Integer operations are common across hardware,
including the latest generation Optimizing Speech Recognition for the Edge of specialized chips.
\item Efficiency. Having all operations as integer means faster execution, and less power consumption.
Furthermore, the use of pre-computed scales means there is no overhead re-computing scales with every inference,
nor quantizing and dequantizing tensors on-the-fly as with the hybrid approach.
\end{itemize}

It is expected that some platform will deviate from the quantization specification due to limitations of hardware capabilities.
The considerations described in this work can help decide where (and how much) deviation is acceptable.

It is also expected that this quantization strategy works for all RNN variants.
Unidirectional-RNN/LSTM and bidirectional-RNN/LSTM have loops on top of LSTM cell
and the quantization strategy described in this work can be directly applied.
For other RNNs such as Gated Recurrent Unit (GRU)~\cite{ChoMGBSB14}, Simple Recurrent Unit (SRU)~\cite{lei2018simple},
the design considerations and building blocks can be use.

\bibliographystyle{unsrt}
\bibliography{references}

\pagebreak
\appendix
\section{APPENDIX}
\begin{table*}[ht!]
\centering
\hspace*{-2.0cm}\begin{tabular}{|l|l|c|c|c|c|c|c|c|c|c|} \hline
tensor & bits & \multicolumn{8}{c|}{scale for each lstm variant} \\\hline
& & No LN & No LN & No LN & No LN & Yes LN & Yes LN & Yes LN & Yes LN \\
& & No Proj & No Proj & Yes Proj & Yes Proj & No Proj & No Proj & Yes Proj & Yes Proj \\
& & No PH & Yes PH & No PH & Yes PH & No PH & Yes PH & No PH & Yes PH \\\hline
$x$ & 8 & $\frac{range}{255}$ & $\frac{range}{255}$ &$\frac{range}{255}$ &$\frac{range}{255}$ &$\frac{range}{255}$ &$\frac{range}{255}$ &$\frac{range}{255}$ &$\frac{range}{255}$   \\\hline
$W_i$\dag & 8 & $\frac{max}{127}$ & $\frac{max}{127}$ &$\frac{max}{127}$ &$\frac{max}{127}$ &$\frac{max}{127}$ &$\frac{max}{127}$ &$\frac{max}{127}$ &$\frac{max}{127}$   \\\hline
$W_f$ & 8 & $\frac{max}{127}$ & $\frac{max}{127}$ &$\frac{max}{127}$ &$\frac{max}{127}$ &$\frac{max}{127}$ &$\frac{max}{127}$ &$\frac{max}{127}$ &$\frac{max}{127}$   \\\hline
$W_z$ & 8 & $\frac{max}{127}$ & $\frac{max}{127}$ &$\frac{max}{127}$ &$\frac{max}{127}$ &$\frac{max}{127}$ &$\frac{max}{127}$ &$\frac{max}{127}$ &$\frac{max}{127}$   \\\hline
$W_o$ & 8 & $\frac{max}{127}$ & $\frac{max}{127}$ &$\frac{max}{127}$ &$\frac{max}{127}$ &$\frac{max}{127}$ &$\frac{max}{127}$ &$\frac{max}{127}$ &$\frac{max}{127}$   \\\hline
$R_i$\dag & 8 & $\frac{max}{127}$ & $\frac{max}{127}$ &$\frac{max}{127}$ &$\frac{max}{127}$ &$\frac{max}{127}$ &$\frac{max}{127}$ &$\frac{max}{127}$ &$\frac{max}{127}$   \\\hline
$R_f$ & 8 & $\frac{max}{127}$ & $\frac{max}{127}$ &$\frac{max}{127}$ &$\frac{max}{127}$ &$\frac{max}{127}$ &$\frac{max}{127}$ &$\frac{max}{127}$ &$\frac{max}{127}$   \\\hline
$R_z$ & 8 & $\frac{max}{127}$ & $\frac{max}{127}$ &$\frac{max}{127}$ &$\frac{max}{127}$ &$\frac{max}{127}$ &$\frac{max}{127}$ &$\frac{max}{127}$ &$\frac{max}{127}$   \\\hline
$R_o$ & 8 & $\frac{max}{127}$ & $\frac{max}{127}$ &$\frac{max}{127}$ &$\frac{max}{127}$ &$\frac{max}{127}$ &$\frac{max}{127}$ &$\frac{max}{127}$ &$\frac{max}{127}$   \\\hline
$P_i$ & 16 & \--- & $\frac{max}{32767}$ & \--- & $\frac{max}{32767}$& \--- & $\frac{max}{32767}$& \--- & $\frac{max}{32767}$  \\\hline
$P_f$ & 16 & \--- & $\frac{max}{32767}$ & \--- & $\frac{max}{32767}$& \--- & $\frac{max}{32767}$& \--- & $\frac{max}{32767}$  \\\hline
$P_o$ & 16 & \--- & $\frac{max}{32767}$ & \--- & $\frac{max}{32767}$& \--- & $\frac{max}{32767}$& \--- & $\frac{max}{32767}$  \\\hline
$b_i$\dag & 32 & $h \times R_i$ & $h \times R_i$ & $h \times R_i$ & $h \times R_i$ & $L_{i} \times 2^{-10}$ & $L_{i} \times 2^{-10} $ & $L_{i} \times 2^{-10}$ & $L_{i} \times 2^{-10}$ \\\hline
$b_f$ & 32 & $h \times R_f$ & $h \times R_f$ & $h \times R_f$ & $h \times R_f$ & $L_{f} \times 2^{-10}$ & $L_{f} \times 2^{-10}$ & $L_{f} \times 2^{-10}$ & $L_{f} \times 2^{-10}$ \\\hline
$b_z$ & 32 & $h \times R_z$ & $h \times R_z$ & $h \times R_z$ & $h \times R_z$ & $L_{z} \times 2^{-10}$ & $L_{z} \times 2^{-10}$ & $L_{z} \times 2^{-10}$ & $L_{z} \times 2^{-10}$ \\\hline
$b_o$ & 32 & $h \times R_o$ & $h \times R_o$ & $h \times R_o$ & $h \times R_o$ & $L_{o} \times 2^{-10}$ & $L_{o} \times 2^{-10}$ & $L_{o} \times 2^{-10}$ & $L_{o} \times 2^{-10}$ \\\hline
$W_{proj}$ & 8 & \--- & \---  &$\frac{max}{127}$ &$\frac{max}{127}$ & \--- & \--- &$\frac{max}{127}$ &$\frac{max}{127}$   \\\hline
$b_{proj}$ & 32 & \--- & \---  & $W_{proj} \times m$ & $W_{proj} \times m$ & \--- & \--- &$W_{proj} \times m$ & $W_{proj} \times m$ \\\hline
$h$ & 8 & $\frac{range}{255}$ & $\frac{range}{255}$ & $\frac{range}{255}$ & $\frac{range}{255}$ & $\frac{range}{255}$ & $\frac{range}{255}$ & $\frac{range}{255}$ & $\frac{range}{255}$ \\\hline
$c$ & 16 & $\frac{POT(max)}{32768}$ & $\frac{POT(max)}{32768}$& $\frac{POT(max)}{32768}$& $\frac{POT(max)}{32768}$& $\frac{POT(max)}{32768}$& $\frac{POT(max)}{32768}$& $\frac{POT(max)}{32768}$& $\frac{POT(max)}{32768}$ \\\hline
$L_i$\dag & 16 & \--- & \---  & \--- & \---  &$\frac{max}{32767}$&$\frac{max}{32767}$&$\frac{max}{32767}$&$\frac{max}{32767}$  \\\hline
$L_f$ & 16 & \--- & \---  & \--- & \---  &$\frac{max}{32767}$&$\frac{max}{32767}$&$\frac{max}{32767}$&$\frac{max}{32767}$  \\\hline
$L_z$ & 16 & \--- & \---  & \--- & \---  &$\frac{max}{32767}$&$\frac{max}{32767}$&$\frac{max}{32767}$&$\frac{max}{32767}$  \\\hline
$L_o$ & 16 & \--- & \---  & \--- & \---  &$\frac{max}{32767}$&$\frac{max}{32767}$&$\frac{max}{32767}$&$\frac{max}{32767}$  \\\hline
$g_i$\dag & 16 & \--- & \---  & \--- & \---  &$\frac{max}{32767}$&$\frac{max}{32767}$&$\frac{max}{32767}$&$\frac{max}{32767}$  \\\hline
$g_f$ & 16 & \--- & \---  & \--- & \---  &$\frac{max}{32767}$&$\frac{max}{32767}$&$\frac{max}{32767}$&$\frac{max}{32767}$  \\\hline
$g_z$ & 16 & \--- & \---  & \--- & \---  &$\frac{max}{32767}$&$\frac{max}{32767}$&$\frac{max}{32767}$&$\frac{max}{32767}$  \\\hline
$g_o$ & 16 & \--- & \---  & \--- & \---  &$\frac{max}{32767}$&$\frac{max}{32767}$&$\frac{max}{32767}$&$\frac{max}{32767}$  \\\hline
$m$ & 8 & \--- & \---  &$\frac{range}{255}$ & $\frac{range}{255}$  & \--- & \---  &$\frac{range}{255}$ & $\frac{range}{255}$  \\\hline

\hline
\end{tabular}
\rule{0in}{1.2em}$^\dag$\scriptsize Rows becomes invalid when CIFG is true; \\
\caption{
The quantization recipe. $max$ is $max(|x_{i}|)$;
$range$ is $max(|x_{i}|) - min(|x_{i}|)$;
“$POT(x)$” is $x$ extended to power of two.
For variants, "LN" means layer normalization, "proj" means projection and "PH" means peephole.
There are scales that are derived from other scales. For example, when there is no layer normlization,
the input bias $b_{i}$ is quantized with the scale $s$ which is the production between the scale for recurrent activation $r$ and recurrent weight $R_{i}$.
Note that tensor cell state $c$ has power of two scales so the denominator is 32768 instead of 32767.
$g_i$, $g_f$, $g_z$, $g_o$ are the gate matrix multiplication output $W x + R h + P \odot c$ for the 4 gates.
}
\label{tab:recipe}
\end{table*}

\begin{figure}[t!]
\centering
\begin{tikzpicture}[node distance = 1.5cm, auto]
    \node [tensor] (w_x) {$W$};
    \node [tensor, right of=w_x] (input_x) {$x$};
    \node [tensor, right of=input_x] (w_h) {$R$};
    \node [tensor, right of=w_h] (input_h) {$h$};
    \node [tensor, right of=input_h] (bias) {$b$};
    \node [tensor, right of=bias] (w_cell) {$P$};
    \node [tensor, right of=w_cell] (cell) {$c$};
    \coordinate (x_center) at ($(w_x)!0.5!(input_x)$);
    \coordinate (h_center) at ($(w_h)!0.5!(input_h)$);
    \coordinate (c_center) at ($(w_cell)!0.5!(cell)$);
    \node [op, below of=x_center] (mul_1) {$\times$};
    \node [op, below of=h_center] (mul_2) {$\times$};
    \node [op, below of=c_center] (mul_3) {$\odot$};
    \node [op, below of=mul_2] (add) {$+$};
    \node [tensor, below of=add] (output) {$y$};

    \path [line] (w_x) -- (mul_1);
    \path [line] (input_x) -- (mul_1);
    \path [line] (w_h) -- (mul_2);
    \path [line] (input_h) -- (mul_2);
    \path [line] (w_cell) -- (mul_3);
    \path [line] (cell) -- (mul_3);
    \path [line] (mul_1) -- (add);
    \path [line] (mul_2) -- (add);
    \path [line] (mul_3) -- (add);
    \path [line] (bias) -- (add);
    \path [line] (add) -- (output);
\end{tikzpicture}
\caption{Gates calculation without layer normalization in original float graph. Cell gate does not have $P$ and $c$.}
\label{fig:gate_no_ln_float}
\end{figure}
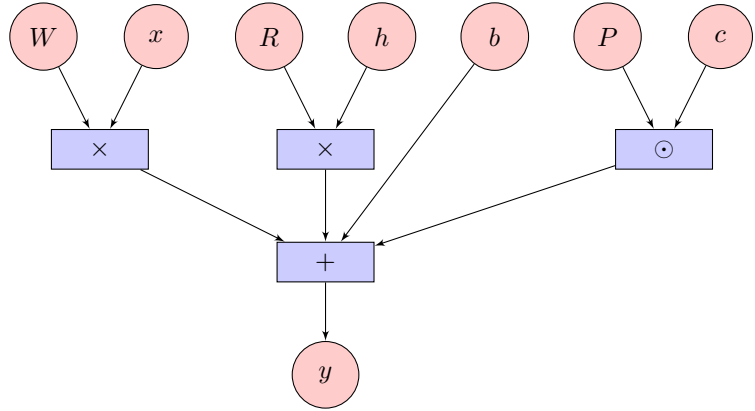

\begin{figure}[t!]
\centering
\begin{tikzpicture}[node distance = 1.5cm, auto]
    \node [tensor] (w_x) {8};
    \node [tensor, right of=w_x] (input_x) {8,8};
    \node [tensor, right of=input_x] (w_h) {8};
    \node [tensor, right of=w_h] (input_h) {8,8};
    \node [tensor, right of=input_h] (bias) {32};
    \node [tensor, right of=bias] (w_cell) {16};
    \node [tensor, right of=w_cell] (cell) {$Q_{m.15-m}$};
    \coordinate (x_center) at ($(w_x)!0.5!(input_x)$);
    \coordinate (h_center) at ($(w_h)!0.5!(input_h)$);
    \coordinate (c_center) at ($(w_cell)!0.5!(cell)$);
    \node [op, below of=x_center] (mul_1) {$\times$};
    \node [op, below of=h_center] (mul_2) {$\times$};
    \node [op, below of=c_center] (mul_3) {$\odot$};
    \node [op, below of=mul_2] (add) {$+$};
    \node [tensor, below of=add] (output) {$Q_{3.12}$};

    \path [line] (w_x) -- (mul_1);
    \path [line] (input_x) -- (mul_1);
    \path [line] (w_h) -- (mul_2);
    \path [line] (input_h) -- (mul_2);
    \path [line] (w_cell) -- (mul_3);
    \path [line] (cell) -- (mul_3);
    \path [line] (mul_1) -- (add);
    \path [line] (mul_2) -- (add);
    \path [line] (mul_3) -- (add);
    \path [line] (bias) -- (add);
    \path [line] (add) -- (output);
\end{tikzpicture}
\caption{Quantized gates calculation without layer normalization. Cell gate does not have $P$ and $c$. The (8, 8) is asymmetrically quantized input and its zero point.}
\label{fig:gate_no_ln_quant}
\end{figure}
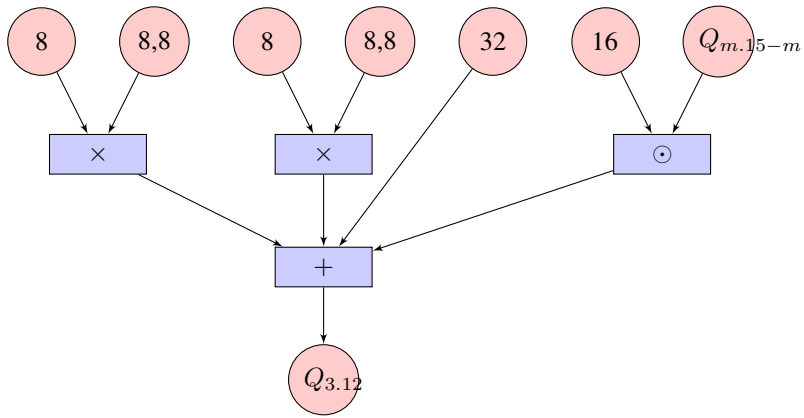

\begin{figure}[t!]
\centering
\begin{tikzpicture}[node distance = 1.5cm, auto]
    \node [tensor] (w_x) {8};
    \node [tensor, right of=w_x] (input_x) {8,8};
    \node [tensor, right of=input_x] (w_h) {8};
    \node [tensor, right of=w_h] (input_h) {8,8};
    \node [tensor, right of=input_h] (w_cell) {16};
    \node [tensor, right of=w_cell] (cell) {$Q_{m.15-m}$};
    \coordinate (x_center) at ($(w_x)!0.5!(input_x)$);
    \coordinate (h_center) at ($(w_h)!0.5!(input_h)$);
    \coordinate (c_center) at ($(w_cell)!0.5!(cell)$);
    \node [op, below of=x_center] (mul_1) {$\times$};
    \node [op, below of=h_center] (mul_2) {$\times$};
    \node [op, below of=c_center] (mul_3) {$\odot$};
    \node [op, below of=mul_2] (add) {$+$};
    \node [tensor, below of=add] (output) {16};

    \path [line] (w_x) -- (mul_1);
    \path [line] (input_x) -- (mul_1);
    \path [line] (w_h) -- (mul_2);
    \path [line] (input_h) -- (mul_2);
    \path [line] (w_cell) -- (mul_3);
    \path [line] (cell) -- (mul_3);
    \path [line] (mul_1) -- (add);
    \path [line] (mul_2) -- (add);
    \path [line] (mul_3) -- (add);
    \path [line] (add) -- (output);
\end{tikzpicture}
\caption{Quantized gates calculation with layer normalization. Cell gate does not have $P$ and $c$. The (8, 8) is asymmetrically quantized input and its zero point.}
\label{fig:gate_ln_quant}
\end{figure}
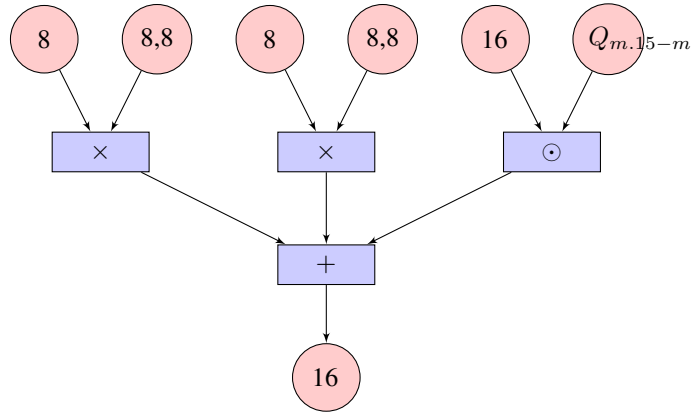

\begin{figure}[t!]
\centering
\begin{tikzpicture}[node distance = 1.5cm, auto]
    \node [tensor] (x) {16};
    \node [op, below of=x] (sum) {sum/div};
    \node [tensor, below of=sum] (x_bar) {32};
    \node [op, right of=x_bar] (stddev) {stddev};
    \node [op, below of=x_bar] (add1) {$+$};
    \node [tensor, below of=add1] (x_0) {16};
    \node [op, below of=x_0] (rescale) {$\times 2^{10}$};
    \node [tensor, below of=rescale] (x_1) {32};
    \node [tensor, right of=x_1] (sigma) {32};
    \node [op, below of=x_1] (mul1) {$\odot$};
    \node [tensor, right of=mul1] (ln_coef) {16};
    \node [tensor, below of=mul1] (x_3) {64};
    \node [op, below of=x_3] (add2) {$+$};
    \node [tensor, right of=add2] (bias) {32};
    \node [tensor, below of=add2] (after_bias) {64};
    \node [op, below of=after_bias] (rescale2) {rescale};
    \node [tensor, below of=rescale2] (output) {$Q_{3.12}$};

    \path [line] (x) -- (sum);
    \path [line] (sum) -- (x_bar);
    \path [line] (x) edge [bend left=30] (stddev);
    \path [line] (x_bar) -- (add1);
    \path [line] (x_bar) -- (stddev);
    \path [line] (stddev) -- (sigma);
    \path [line] (x) edge [bend right=50] (add1);
    \path [line] (add1) -- (x_0);
    \path [line] (x_0) -- (rescale);
    \path [line] (rescale) -- (x_1);
    \path [line] (x_1) -- (mul1);
    \path [line] (sigma) -- (mul1);
    \path [line] (ln_coef) -- (mul1);
    \path [line] (mul1) -- (x_3);
    \path [line] (x_3) -- (add2);
    \path [line] (bias) -- (add2);
    \path [line] (add2) -- (after_bias);
    \path [line] (after_bias) -- (rescale2);
    \path [line] (rescale2) -- (output);
\end{tikzpicture}
\caption{Layer normalization integer execution.} \label{fig:ln_execution}
\end{figure}
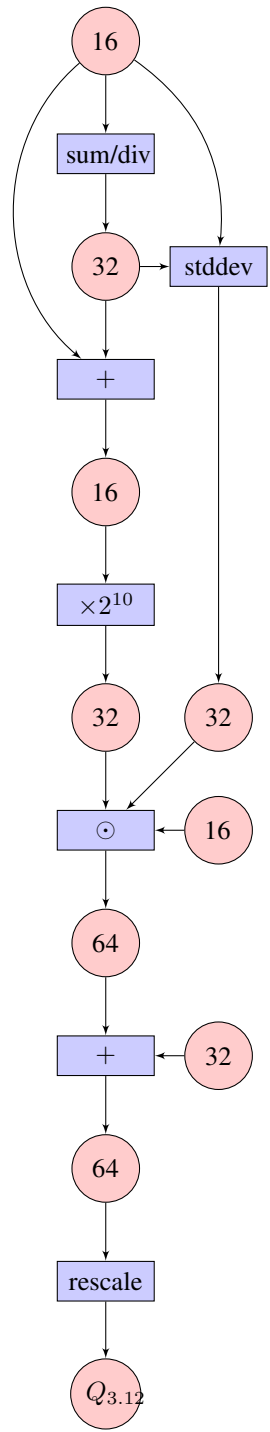

\begin{figure}[t!]
\centering
\begin{tikzpicture}[node distance = 1.5cm, auto]
    \node [tensor] (weight) {$W$};
    \node [tensor, right of=weight] (input) {$m$};
    \node [tensor, right of=input] (bias) {$b$};
    \coordinate (h_center) at ($(weight)!0.5!(input)$);
    \node [op, below of=h_center] (mul) {$\times$};
    \node [op, below of=mul] (add) {$+$};
    \node [tensor, below of=add] (output) {$h$};

    \path [line] (weight) -- (mul);
    \path [line] (input) -- (mul);
    \path [line] (mul) -- (add);
    \path [line] (bias) -- (add);
    \path [line] (add) -- (output);
\end{tikzpicture}
\caption{Float execution for projection.}
\label{fig:projection_float}
\end{figure}
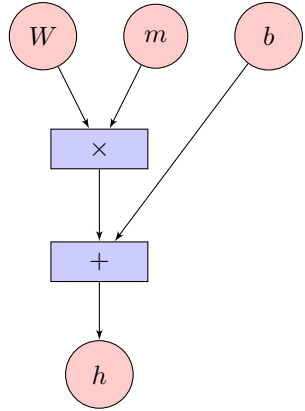

\end{document}